\documentclass[11pt]{article}

\usepackage[final]{acl}

\usepackage{times}
\usepackage{latexsym}

\usepackage[T1]{fontenc}

\usepackage[utf8]{inputenc}

\usepackage{microtype}

\usepackage{inconsolata}

\usepackage{graphicx}
\usepackage{booktabs}
\usepackage{url}
\usepackage{tikz}
\usepackage{xcolor}
\usetikzlibrary{positioning,arrows.meta,shapes.geometric,shadings,calc,fit,backgrounds}
\usepackage{amsmath}
\usepackage{paralist}
\usepackage{tabularx}
\usepackage{multirow}
\newcommand{\zwReviewMode}{0}
\ifnum\zwReviewMode=1
  \newcommand{\zw}[1]{\textcolor{red}{\textbf{[ZW: #1]}}}
  \newcommand{\zwbox}[1]{%
    \medskip\noindent
    \fcolorbox{red}{red!8}{%
      \parbox{0.95\columnwidth}{%
        \color{red}\small\textbf{ZW:} #1%
      }%
    }\medskip%
  }
  \newcommand{\todoz}[1]{\colorbox{yellow}{\textbf{TODO:} #1}}
  
\else
  \newcommand{\zw}[1]{}
  \newcommand{\zwbox}[1]{}
  \newcommand{\todoz}[1]{}
  
\fi

\title{SIC-Agents: Benchmarking and Building an Adaptive Simulator for
\\ Pediatric Serious Illness Communication Training}

\author{
  \textbf{Zihan Wang\textsuperscript{1,2}},
  \textbf{Anita Marie Slominska\textsuperscript{3}},
  \textbf{Rennie Bimman\textsuperscript{4}},
\\
  \textbf{Elizabeth Di Flumeri\textsuperscript{5}},
  \textbf{Amanda Mayappo-Neeposh\textsuperscript{5}},
  \textbf{Conall Francoeur\textsuperscript{5,6}},
\\
  \textbf{Tamara Ellen Carver\textsuperscript{3,7}},
  \textbf{Xiao-Wen Chang\textsuperscript{1}},
  \textbf{Doina Precup\textsuperscript{1,2}},
\\
  \textbf{Esin Darici Haritaoglu\textsuperscript{8,\dag}},
  \textbf{Ismail Haritaoglu\textsuperscript{8,\dag}},
  \textbf{Akshatha Arodi\textsuperscript{2,\dag}},
  \textbf{Naomi Goloff\textsuperscript{3,5,\dag}}
\\
\\
  \textsuperscript{1}School of Computer Science, McGill University,
  \textsuperscript{2}Mila -- Quebec AI Institute,
\\
  \textsuperscript{3}Institute of Health Sciences Education, McGill University,
\\
  \textsuperscript{4}School of Social Work, McGill University,
\\
  \textsuperscript{5}Department of Pediatrics, McGill University,
\\
  \textsuperscript{6}Montreal Children's Hospital, McGill University Health Centre,
\\
  \textsuperscript{7}Steinberg Centre for Simulation and Interactive Learning, McGill University,
\\
  \textsuperscript{8}Linarite AI
\\
  \small{
    \textsuperscript{\dag}Equal supervision.
  }
\\
  \small{
    \textbf{Correspondence:} \href{mailto:zihan.wang@mila.quebec}{zihan.wang@mila.quebec}
  }
}

\begin{document}
\maketitle

\begin{abstract}
Pediatric serious illness communication (SIC) is critically important, yet scalable communication
training for clinicians remains limited. Compared with other dialogue simulation settings, pediatric SIC poses additional challenges, including multi-party interactions, response to parental distress and strong dependence on feedback dynamics. Existing LLM-based simulators optimize generic dialogue quality rather
than curriculum-contingent behavior required for effective SIC training. In collaboration with educators and pediatric clinicians, we introduce the first benchmark suite and simulation framework tailored to pediatric SIC training. Our benchmarks, PitfallBench and
DialogueBench, evaluate simulators both at the turn-level and across full dialogues.
We further propose SIC-Agents, a self-improving framework that generates a clinician-editable skill
document to guide simulator behavior. Our experiments show that SIC-Agents outperforms
static expert prompting.
To support future research, we release our
benchmarks\footnote{\url{https://github.com/Beikewzh/sic-benchmarks}}
for parent simulation in pediatric SIC.
\end{abstract}

\section{Introduction}
\label{sec:intro}

When baby O was transferred to a new hospital, the medical team
informed her family that brain imaging showed ``evidence'' of a
terminal genetic disease. Medical facts were delivered, but the
clinicians failed to acknowledge the emotional impact or meaning,
leaving the family terrified, feeling pushed aside, and wondering
whether their baby was going to die.\footnote{Based on a real parent experience contributed by a parent-partner coauthor; full vignette in
Appendix~\ref{app:vignette}.} 
Experiences like this illustrate that
serious illness communication\footnote{SIC refers to the communication skills required to care for patients and families facing serious or life-limiting illness \citep{feudtner2007collaborative, bernacki2014communication, jacobsen2022shifting, henderson2024goals}.} (SIC) requires more than the delivery
of medical information alone. Yet, despite the critical importance of
these communication skills, pediatric clinicians consistently report
feeling underprepared \citep{file2014pho, wolfe2016badnews,
rossfeld2018selfassess, goloff2025ppc, dean2019moral}.

\begin{figure*}[!t]
\centering
\includegraphics[width=\textwidth]{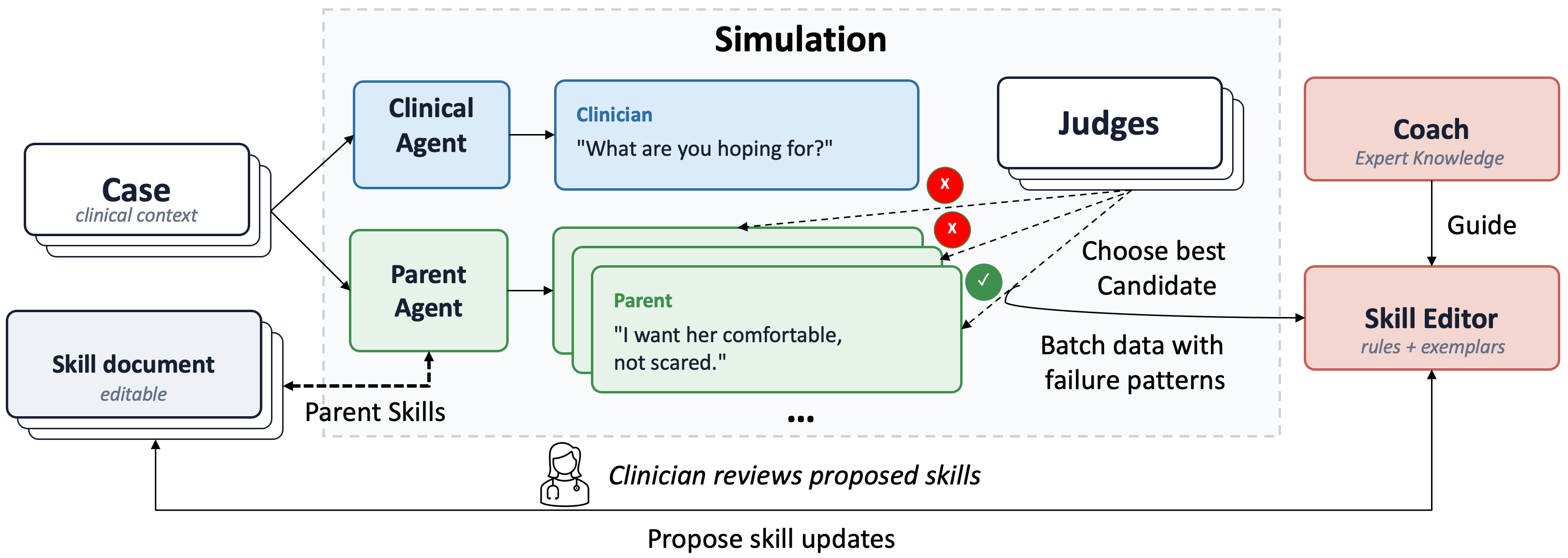}
\caption{SIC-Agents self-improvement loop. A parent agent generates
candidate responses conditioned on the case, dialogue history, and
editable skill document. LLM judges select responses and provide
feedback, which a Skill Editor uses to revise the skill document under
curriculum guidance from a Coach agent. Clinicians and human instructors can audit and
edit both the skill document and revisions.}
\label{fig:sicagents_arch}
\end{figure*}

Current SIC training relies largely on resource-intensive,
difficult-to-scale standardized patient (SP) simulations, where an
actor portrays a patient or parent and an observer provides feedback
\citep{cleland2009simulated, back2007oncotalk}. While large language
model (LLM) simulators offer a promising scalable alternative by
allowing clinicians to engage in repeated, on-demand simulated
conversations, and have been explored for psychotherapy, diagnostic
interviewing, and adult palliative care communication
\citep{tu2024amie, wang2024patientpsi, wang2024pallm, haut2025sophie,
sehgal2025pal}, pediatric SIC differs significantly from these
settings. In pediatrics, clinicians must navigate complex multi-party
communication, respond to parental distress, communicate
prognostic uncertainty, and support surrogate decision-making while
maintaining trust during moments of vulnerability
\citep{vanbreemen2020casereview, henderson2024goals}.

This pediatric distinction exposes an important gap for the natural
language processing (NLP) community. Existing medical dialogue
simulators primarily emphasize information exchange, role adherence, or
task completion. Pediatric SIC is a
feedback-dependent interaction: clinician language continuously shapes
parental emotion, trust, and engagement, which in turn alters
subsequent conversational behavior \citep{cleland2009simulated,
lewis2017aspe}. Crucially, this feedback-dependent interaction is
important not only for accurately modeling clinical communication, but
also for the underlying educational learning model itself; acquiring
durable communication skills requires immediate feedback and
iterative, deliberate practice \citep{deoliveira2024rcdp,ericsson2008deliberate}. Effective simulation environments must
therefore model not only clinical content, but also dynamically
evolving interpersonal and emotional states across the course of a
conversation. Despite their importance, such adaptive relational
dynamics remain largely underexplored in NLP due to limited data availability, privacy constraints, and the clinician involvement required for development. More broadly, pediatric SIC introduces an
important research area for adaptive dialogue systems operating in
emotionally sensitive and multi-party settings.

To address this gap, we collaborated with educators and pediatric clinicians to develop benchmarks and simulation methods aligned with clinical training needs. We introduce two benchmarks
and \emph{SIC-Agents}, a self-improving parent simulator for pediatric SIC. We study whether a data-scarce simulator can improve through self-generated practice while remaining aligned with
clinician-authored curriculum guidance. Unlike latent policy
optimization approaches, SIC-Agents maintains a clinician-readable skill document that enables auditing, correction,
and curriculum control. Because
conversational quality depends not only on factual correctness but also
on relational appropriateness, emotional responsiveness, and adaptation
across conversational turns, evaluation is particularly challenging. We therefore propose benchmarks and clinician-defined evaluation dimensions to measure simulator behavior changes
in response to supportive communication skills and common pitfalls.
Together, the benchmarks and simulation framework position pediatric SIC as
a testbed for emotionally grounded, curriculum-conditioned dialogue
modeling in high-stakes settings.

Developed with educator and clinician co-authors, this paper makes three contributions:
\begin{compactenum}
\item \textbf{Evaluation Benchmarks and Metrics.}
We introduce \textsc{PitfallBench} and \textsc{DialogueBench},
evaluation benchmarks for pediatric SIC, at the turn level and across full dialogues, respectively. The benchmarks compare supportive and suboptimal clinician communication strategies in matched scenarios. Through iterative expert refinement, we additionally
define evaluation dimensions for emotional responsiveness, relational
adaptation, and conversational grounding
(\S\ref{sec:dataset}).

\item \textbf{Inspectable Agentic Simulation.} We develop
\emph{SIC-Agents}, a multi-agent simulation framework that
self-improves a clinician-editable skill document
while using curriculum guidance to reduce drift, i.e.,
deviations from medically and behaviorally appropriate parent responses
(\S\ref{sec:framework},~\S\ref{sec:experiments}) (Figure ~\ref{fig:sicagents_arch}).
\item \textbf{Training Infrastructure for Deliberate Practice.} We
integrate the simulator into a rapid-cycle deliberate-practice web-interface for SIC training, to support scalable evaluation
and clinical education.
\end{compactenum}

\section{Background and Related Work}
\label{sec:background}

\subsection{Identifying Learning Opportunities and Conversational Flow}

Rather than strictly evaluating clinical competency, SP encounters are
designed to track the flow of the conversation and identify critical
learning opportunities where faculty can provide formative feedback
\citep{cleland2009simulated, back2007oncotalk, paladino2019sicp}. In clinical
education, learners are assessed on their ability to execute specific,
clinically meaningful moves based on established serious illness
frameworks \citep{ko2020sicex}. To operationalize these moves for
targeted feedback, this work isolates the \emph{trigger turn}: the
exact point at which the learner commits one of these meaningful
moves, acting as either a communication pitfall or its skilled
alternative. Identifying these turns allows faculty to provide
immediate feedback and, crucially, requires the simulated parent's
response on the immediately following turn to dynamically adapt to
what the clinician just said.

While established guides outline the core communication skills
required for these conversations \citep{daubman2021best,
vanbreemen2024sicgpeds}, effectively simulating them requires mapping
where learners most commonly struggle \citep{meyer2009difficult}.
Drawing on extensive faculty teaching experience within both
pediatric and adult SIC programs, we empirically derived a taxonomy of
ten common learner mistakes (pitfalls) distributed across two primary
conversation phases: the ability to effectively share and support
medical information, and the ability to explore goals and values and
align the medical plan. For each pitfall, this expert-derived
framework provides the core skill the learner forgets alongside the
skilled-clinician alternative. Several of these observed pitfalls are
uniquely tied to the pediatric relational dynamic, such as asking
which treatments parents want rather than what they hope for, or
framing what the clinical team will not do before framing what it will
do \citep{vanbreemen2024sicgpeds}.

While the trigger turn provides a localized learning opportunity, the
overarching conversational flow must also be maintained. A second key
metric is the \emph{across-session} judgment of whether the whole
encounter stayed grounded and topically consistent over many turns.
While relevant to human SP encounters, this metric is primarily used
here to evaluate the difference between the output of an AI simulator
compared to what a human actor would do, as an artificial dialogue can
easily become disjointed even when individual turns are locally
appropriate. A successful parent simulator must therefore be evaluated
against both its contingent adaptation at the trigger turn and its
ability to maintain a grounded, human-like dialogue across the entire
encounter.

\subsection{The Gap in Existing Conversational Agents}

Communication-skill training is bottlenecked by scalable practice:
durable acquisition requires repeated feedback-driven interaction that
one-off SP workshops cannot provide
\citep{ericsson2008deliberate, hunt2014rcdp, file2014pho}. This has
motivated LLM-based patient simulators built on earlier
LLM-as-character work
\citep{park2023generative, shao2023characterllm}. However, prior
systems are evaluated mainly on generic realism metrics; none targets
pediatric SIC, evaluates contingent adaptation at curriculum-defined
pitfalls, or represents knowledge as a clinician-editable document.

Addressing this gap requires behavioral evaluation focused on specific
capabilities rather than held-out accuracy
\citep{ribeiro2020beyond}. Similar ideas have recently appeared in
clinical evaluation through MATRIX \citep{lim2025matrix}. We apply this
framework in pediatric SIC, where privacy constraints preclude large
public corpora \citep{singhal2023medpalm, luo2022biogpt}. Accordingly,
the benchmarks in \S\ref{sec:dataset} evaluate contingent adaptation at
trigger turns and grounding across extended dialogue. A comparison with existing clinician-anchored dialogue benchmarks is in Table \ref{tab:bench_comparison}. Prior
self-improving LLM frameworks such as Reflexion
\citep{shinn2023reflexion}, SELF-REFINE
\citep{madaan2023selfrefine}, ACE \citep{zhang2025ace}, and
Constitutional AI \citep{bai2022constitutional} show iterative
behavioral refinement without weight updates. However, generic
self-improvement targets can improve surface dialogue quality while
missing the curriculum-contingent behavior that makes a pediatric SIC
simulator clinically valid. SIC-Agents therefore uses these standard
self-improvement components, but also embeds pediatric SIC
curriculum feedback in a clinician-editable form (\S\ref{sec:framework}).

\section{Datasets}
\label{sec:dataset}

Public pediatric serious-illness-conversation (SIC) dialogue data are effectively unavailable because of clinical confidentiality constraints \citep{singhal2023medpalm, luo2022biogpt}. Consequently, prior work has lacked both (i) realistic substrates for training pediatric-parent simulators and (ii) clinician-anchored evaluation resources for measuring simulator quality. To address this gap, clinician co-authors with extensive experience in SIC training identified two key evaluation targets (\S\ref{sec:background}): \emph{turn-level contingent adaptation} and \emph{dialogue-level grounding}. Accordingly, we release two corresponding resources:

\begin{compactitem}
    \item \textbf{\textsc{PitfallBench}}: a $1{,}000$-item turn-level benchmark that evaluates whether a parent simulator reacts appropriately to clinician communication quality at a \emph{trigger turn}; and
    \item \textbf{\textsc{DialogueBench}}: a $152$-dialogue gold benchmark that evaluates whether a simulator maintains coherent grounding and role consistency across a full multi-turn encounter. Accompanied by SIC-SimCorpus, an $8{,}391$-dialogue silver corpus.
\end{compactitem}

Both resources are constructed in close collaboration with pediatric SIC clinicians with training experience and are anchored in the Serious Illness Conversation Guide for Pediatrics curriculum \citep{daubman2021best, vanbreemen2024sicgpeds}. Unlike benchmarks based primarily on post hoc per-item annotation (Table \ref{tab:bench_comparison}), our resources define validity through an explicit, auditable acceptance criterion jointly developed and refined by clinicians and an independent LLM reviewer.


\begin{table}[t]
\centering
\footnotesize
\setlength{\tabcolsep}{3pt}
\renewcommand{\arraystretch}{1.05}

\begin{tabularx}{\columnwidth}{@{}X l l l@{}}
\toprule
\textbf{Benchmark} & \textbf{Domain} & \textbf{Gold} & \textbf{Expert} \\
\midrule

Patient-$\Psi$ \citep{wang2024patientpsi} & Psychotherapy & 106 & designed \\
PatientSim \citep{kyung2025patientsim}    & General  & 37  & checked \\
AMIE \citep{tu2024amie}                   & Diagnostic        & 149 & rated \\
\midrule

\textbf{PitfallBench (ours)}  & Pediatric SIC & \textbf{1,000} & curriculum\\
\textbf{DialogueBench (ours)} & Pediatric SIC & \textbf{152}   & curriculum\\
\bottomrule
\end{tabularx}

\caption{Comparison with existing clinician-anchored dialogue benchmarks.
\emph{Gold} denotes the number of released evaluation items. Our benchmarks are grounded in an external pediatric SIC curriculum and refined under expert oversight.}
\label{tab:bench_comparison}
\end{table}
\subsection{\textsc{PitfallBench}}
\label{sec:dataset:pitfallbench}

\begin{figure}[!t]
\centering
\includegraphics[width=0.85\columnwidth]{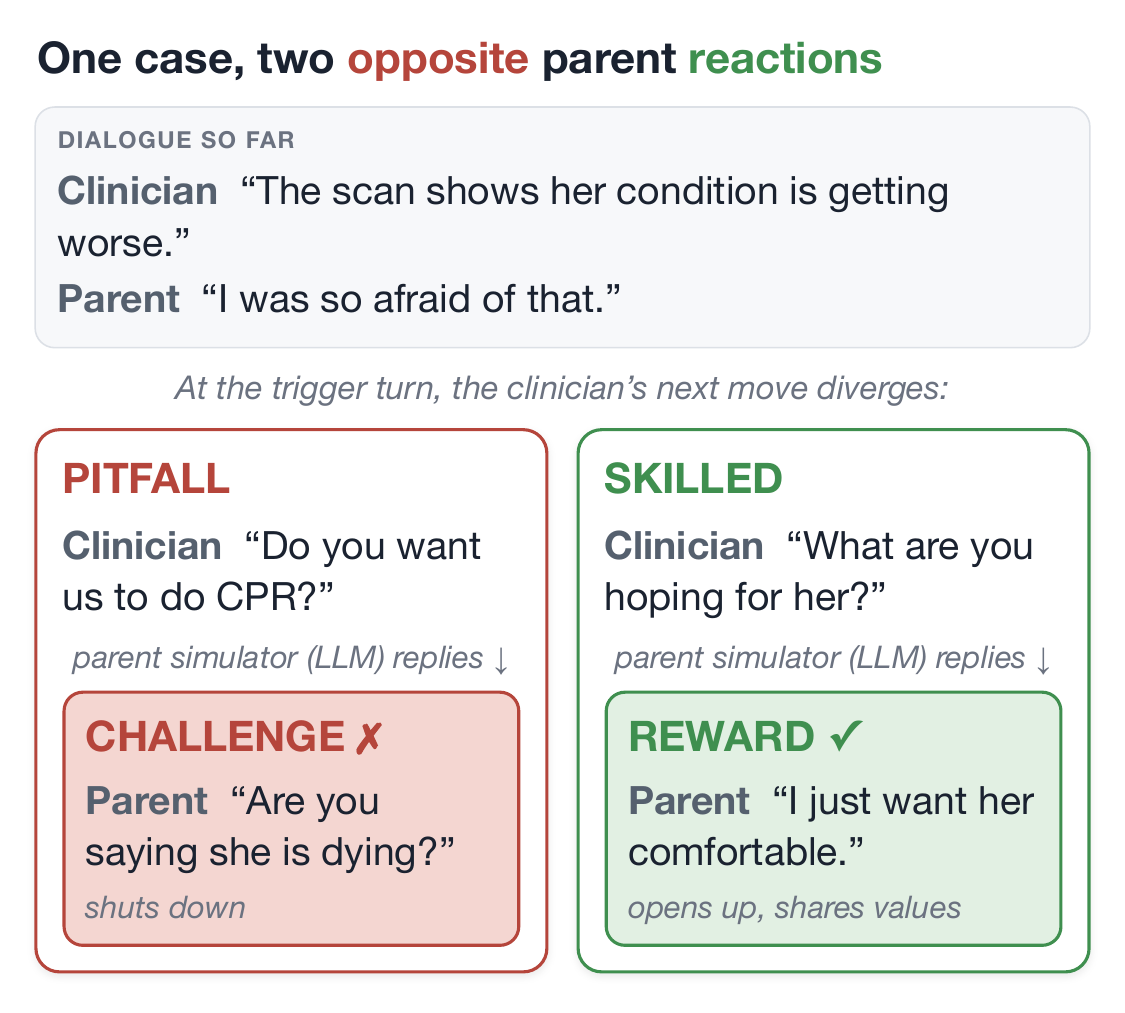}
\caption{PitfallBench uses a paired within-case design. The same
case and dialogue history are instantiated twice at the trigger
turn: once with a curriculum-defined pitfall move and once with
its skilled alternative. The simulator must generate
differentially appropriate parent responses under otherwise
identical conversational conditions.}
\label{fig:pitfallbench_example}
\end{figure}

PitfallBench evaluates contingent adaptation at the level of a single \textit{trigger turn}. The benchmark is motivated by a central pedagogical objective in pediatric SIC training: a clinician utterance that follows best practice should elicit a meaningfully different parent response than a clinically suboptimal utterance under the same conversational context.

\paragraph{Task formulation}

Each \textsc{PitfallBench} item consists of a pediatric clinical case, a partial dialogue history up to a \emph{trigger turn}, a curriculum-defined \emph{pitfall} clinician move, its corresponding \emph{skilled} alternative, and the expected parent reaction associated with each move (Figure \ref{fig:pitfallbench_example}). At evaluation time, the simulator receives the case context, the dialogue history, and one clinician move, and generates the next parent utterance. A judge LLM then classifies the generated response as \texttt{challenge\_signal}, \texttt{reward\_signal}, or \texttt{neither} relative to the curriculum-defined expected reaction. The benchmark therefore operationalizes contingent adaptation as a constrained three-way classification problem over simulator behavior.

\paragraph{Curriculum structure}

PitfallBench is organized around ten common learner pitfalls (Table~\ref{tab:pitfallbench_composition}) compiled by our pediatric SIC clinician co-authors (Appendix~\ref{app:annotators}). Each pitfall is paired with a \emph{curriculum rule}, an expert-authored template that fixes the dialogue state, the suboptimal clinician move, its skilled counterpart, and the \emph{expected reaction label} that becomes the gold class (\texttt{challenge\_signal} for the pitfall, \texttt{reward\_signal} for the skilled alternative; per-pitfall sketches in Appendix Table~\ref{tab:pitfallbench_rules}). The cue--response contingencies these rules encode are documented in expert-annotated recordings of real pediatric care conferences \citep{olszewski2023empathy}, whose observed family responses correspond to the reward and challenge dynamics our items formalize.

\paragraph{Paired evaluation design}

The benchmark uses a paired within-case construction in which both the pitfall move and the skilled alternative are instantiated under the same dialogue state. This design controls for case-level variation and isolates the behavioral effect of the clinician move itself. It also reduces degenerate evaluation strategies: simulators that default to emotional escalation fail the skilled condition, while simulators that produce uniformly neutral responses fail the pitfall condition. Successful performance therefore requires contingent behavioral discrimination between the paired stimuli.

\paragraph{Gold-label construction}

A central methodological contribution of \textsc{PitfallBench} is its criterion-based validation protocol. Rather than relying exclusively on per-item expert annotation, we define validity through an explicit acceptance criterion iteratively calibrated between pediatric SIC clinicians and an independent LLM reviewer from a different model family than the generator.

Candidate items are first generated from curriculum templates, after which the reviewer applies the current written acceptance criterion. Clinicians (Appendix~\ref{app:annotators}) then inspect reviewer decisions on a stratified sample. Disagreements trigger revisions to the criterion and generation prompts rather than manual editing of individual items, and failed items are regenerated and re-screened. Iteration terminates once reviewer--clinician agreement reaches $95\%$ on a stratified validation sample of $50$ items ($5$ per pitfall). The resulting frozen criterion is then applied uniformly across all released items, with every screening decision logged together with the criterion version. This protocol follows the screen-and-regenerate paradigm used in recent synthetic clinical benchmark construction \citep{wang2024patientpsi, kyung2025patientsim} while making the acceptance rule itself an explicit and auditable object.

\paragraph{Dataset composition}
The benchmark contains $100$ underlying pediatric cases stratified across $12$ pediatric care settings, child ages $0$--$18$, multiple disease categories, and diverse family configurations. Each of the $1{,}000$ items embeds a partial dialogue history of up to ten turns (median eight) leading into the trigger turn. An embedding-based diversity audit (Appendix~\ref{app:diversity}) shows broad semantic coverage across clinical scenarios while preserving tight within-case clustering for paired comparisons.

\subsection{\textsc{DialogueBench}}
\label{sec:dataset:synthetic}

While \textsc{PitfallBench} evaluates local contingent adaptation, \textsc{DialogueBench} evaluates dialogue-level grounding across a full multi-turn pediatric SIC encounter.

\paragraph{Substrate generation}

Because no public pediatric SIC corpus exists, we first construct a synthetic dialogue substrate (SIC-SimCorpus) using a factorized, persona-conditioned generation pipeline. The pipeline varies eight LLM generator families, six pediatric disease categories, three trainee proficiency levels, and dialogue lengths ranging from $12$--$28$ turns. Using multiple generator families reduces dependence on any single model distribution and improves stylistic diversity. Full generation statistics are in
Appendix~\ref{app:diversity}.

\begin{table}[!ht]
\centering
\small
\setlength{\tabcolsep}{4pt}

\begin{tabularx}{\columnwidth}{@{}l X X@{}}
\toprule
\textbf{ID} & \textbf{Pitfall} & \textbf{Core SIC skill} \\
\midrule

\multicolumn{3}{@{}l}{\emph{Phase 1: share and support information}} \\

P1 & Missed early emotion cue & Ask for permission \\
P2 & Forgets to assess understanding & Assess understanding \\
P3 & Information too complex & Headline framing \\
P4 & Missed or combined empathy cues & Respond to emotion \\
P5 & Stuck in validation & Move on after empathy \\

\midrule

\multicolumn{3}{@{}l}{\emph{Phase 2: explore goals and align plan}} \\

P6 & Straight to code status & Elicit values first \\
P7 & Single or too-broad value & Multiple specific values \\
P8 & Cannot consolidate or align & Consolidate + recommend \\
P9 & Asks treatments not values & Hopes / worries framing \\
P10 & ``Won't do'' before ``will do'' & Lead with will \\

\bottomrule
\end{tabularx}

\caption{The ten learner pitfalls covered by PitfallBench, grouped by curriculum phase, with the core SIC skill each pitfall compromises.}
\label{tab:pitfallbench_composition}
\end{table}

\paragraph{Gold/silver release structure}

Following the silver-standard methodology established in biomedical NLP \citep{rebholz2010calbc}, we adopt a gold/silver release structure. The Silver tier consists of the full $8{,}391$-dialogue synthetic corpus, released as a finetuning substrate, while the Gold tier consists of the final $152$-dialogue evaluation benchmark reviewed by human experts. Gold-tier selection proceeds in two stages. An automated scoring pass first removes dialogues that fail a fixed threshold on a two-axis rubric measuring Conversational Grounding (responsiveness to the clinician's latest turn) and Topic Consistency (maintenance of a coherent concern). The rubrics are defined by pediatric clinicians. Pediatric SIC clinicians then review the surviving dialogues for clinical realism using the same frozen acceptance criterion applied in \textsc{PitfallBench}. Using a shared criterion ensures that both benchmarks inherit the same operational definition of clinical validity.

\begin{figure}[t]
\centering
\small
\fbox{\parbox{0.92\columnwidth}{%
\textbf{Case:} 7\,yo with progressive neurodegenerative
disease; outpatient palliative-care consultation; mother
present.\\[2pt]
\textbf{Clinician:} I want to make sure we are on the same
page about what the team is most worried about today.\\
\textbf{Parent:} I just keep hoping the new medication will
start to work.\\
\textbf{Clinician:} It is hard to hold both hope and worry at
the same time. Can you tell me what you have been most
worried about lately?\\
\textbf{Parent:} I am worried that she is not going to be
here much longer. (\emph{long pause}) I do not know how to
tell her brother.\\
$[\ldots]$
}}
\caption{Excerpt from a \textsc{DialogueBench}. During evaluation, clinician turns are fixed and the simulator generates the parent turns. Responses are scored on Conversational Grounding and Topic Consistency.}
\label{fig:dialoguebench_sample}
\end{figure}

\paragraph{Evaluation protocol}

During evaluation, the clinician turns from a benchmark dialogue are held fixed as a script. The simulator generates parent responses in place of the original turns, and generated responses are scored by two independent LLM judges on a two-axis $1$--$3$ rubric for \emph{Conversational Grounding} and \emph{Topic Consistency}. The judges are
calibrated against annotations from $11$ pediatric clinicians; see Appendix~\ref{app:rubric}.
Because the clinician script is identical across systems, evaluation is performed under matched conversational conditions, enabling direct item-level comparison between simulators. The original parent turns additionally serve as a reference condition in the matched comparisons reported in \S\ref{sec:experiments}.




\section{SIC-Agents}
\label{sec:framework}

SIC-Agents is a curriculum-conditioned self-improving parent simulation framework (Figure~\ref{fig:sicagents_arch}) designed with two objectives: (1) to generate interactional training evidence without requiring access to private clinical dialogue data, and (2) to preserve interpretability by representing adaptive parent behavior in an editable natural-language skill document rather than in latent model parameters.

The encounter is anchored by a fixed clinician agent guided by the same pediatric SIC curriculum \citep{daubman2021best, vanbreemen2024sicgpeds}. Its role is to
provide a stable learner-side interlocutor that follows the SIC guide
well enough to create coherent practice turns and, in self-improvement runs, to
surface the kinds of clinician moves the parent simulator must react to.


The parent agent is conditioned on the pediatric case, dialogue history, and a mutable skill document \(S\). The skill document serves as an editable behavioral
specification, encoding six expression patterns (Appendix~\ref{app:sicagents}), exemplar utterances,
transition guidance, and constraints on length, role fidelity, emotional
calibration, and off-topic generation.
Because \(S\) is represented in natural language, clinical experts can inspect and
revise it directly, and the parent agent can use the revised document in
subsequent simulation cycles.

In each simulation cycle, the parent agent samples \(K\) candidate responses at every parent turn \(t\) (\(K{=}3\) in our experiments; Appendix~\ref{app:expconfig}). Invalid generations are removed via a hard filter, after which the remaining candidates are evaluated by an LLM-judge ensemble along the dimensions of \textit{Conversational Grounding} and \textit{Topic Consistency} (Appendix~\ref{app:rubric}). The highest-scoring candidate is selected for inclusion in the dialogue, while both selected and rejected candidates are retained as contrastive evidence for the Skill Editor. This process produces a lightweight source of self-generated preference data that exposes both the strengths and failure modes of the current skill document \(S\).

Conditioned on this batch evidence, the Skill Editor proposes localized modifications to \(S\), such as adding exemplars, refining constraints, revising response patterns, or introducing previously missing interaction strategies. Consequently, adaptation occurs over the interpretable skill document rather than the underlying LLM parameters. Across simulation cycles, \(S\) incrementally accumulates interaction-derived behavioral rules and exemplars from simulated practice.

During self-improvement, the Skill Editor receives two complementary feedback streams: Judge-derived dialogue feedback and Coach-derived curriculum feedback. The judge ensemble identifies interaction-level deficiencies, including failures in conversational grounding, topic maintenance, response length, and role fidelity. However, these scores do not capture the pedagogical context of the interaction. The Coach provides this missing curriculum-level supervision by specifying which learner pitfall a clinician behavior instantiates, which SIC skill is being exercised, and what parent reaction should appropriately follow. This information is derived from the same pediatric SIC curriculum and includes the learner pitfall, target SIC skill, and expected parent response trajectory. During evaluation, neither feedback stream is available to the parent agent, which operates solely from the frozen skill document \(S\).

The resulting skill document powers an open-source training platform where learners engage the parent agent in multi-turn interactions with curriculum-grounded feedback. Faculty can route challenging cases back through the Skill Editor, enabling iterative refinement from deployment data. More broadly, SIC-Agents supports confidential, data-scarce domains by improving an interpretable policy via curriculum-guided self-improvement, and extends naturally to other feedback-intensive settings such as mental-health counseling, crisis intervention, and social-work training.


\section{Experiments}
\label{sec:experiments}

We conduct experiments on SIC-Agents as a feedback-distillation framework operating over a frozen skill document \(S\). At evaluation time, the parent agent receives no external feedback and generates responses solely conditioned on \(S\). We investigate whether static prompting is sufficient, whether self-improvement without curriculum supervision is adequate, and whether curriculum-level supervision is necessary to preserve clinically contingent behavior.

We evaluate on \textsc{PitfallBench} (\S\ref{sec:dataset:pitfallbench}), which
measures whether the parent agent reacts appropriately to a curriculum-defined
pitfall or skilled clinician move at a matched trigger turn, and on Gold
\textsc{DialogueBench} (\S\ref{sec:dataset:synthetic}), which measures
dialogue-level grounding and topic consistency. Our primary outcome is
\textsc{PitfallBench} accuracy, because it tests the curriculum-contingent
behavior that generic dialogue feedback does not label. We also report
\textsc{DialogueBench} Grounding and a $0$--$1$ composite that averages
PitfallBench accuracy with grounding rescaled to $0$--$1$ as secondary
summaries.
Because the generation loop itself uses grounding/topic judges,
DialogueBench is partly aligned with the generic feedback stream. We
therefore interpret it as a check that dialogue-quality feedback was
retained in the skill document, not as an independent proof of clinical
alignment.
Table~\ref{tab:main_headline} gives the headline results; full
per-condition breakdowns are in Appendix~\ref{app:results}.

\paragraph{Experimental settings.}
We compare five deployable conditions that isolate where the curriculum
enters the system:
\begin{compactitem}
\item \emph{No loop}: a deliberately minimal parent seed; an expert
prompt that places expert-written SIC principles directly in the
parent's context; and a full-taxonomy prompt that further adds the
explicit P1--P10 taxonomy and expected reactions to the expert prompt
(Appendix~\ref{app:expconfig}).
\item \emph{Self-improvement}: two systems begin from the same minimal
seed and run five edit cycles. The \emph{no-curriculum} ablation keeps candidate sampling, a panel of judges, and
the Skill Editor, but removes the Coach signal. The full system
adds the Coach signal.
\end{compactitem}

\paragraph{Curriculum-signal isolation.}
Because the curriculum signal derives from the same pitfall taxonomy used
to construct PitfallBench, we restrict it to self-improvement. During
training, the clinician agent uses the same curriculum-guided configuration
across conditions, so the ablation isolates the Coach and Skill Editor
rather than the simulated encounter itself (Appendix~\ref{app:expconfig}).
During benchmark evaluation, clinician turns are fixed scripts from
PitfallBench or DialogueBench, and all deployable parent agents run from
the frozen skill document only; the parent never sees the item label or
expected reaction. An oracle variant that reveals the signal at
inference is reported only as an upper bound in Appendix~\ref{app:results}.

\begin{table}[t]
\centering
\small
\setlength{\tabcolsep}{5pt}

\begin{tabularx}{\columnwidth}{@{}Xccc@{}}
\toprule
\textbf{Condition} & \textbf{Primary} & \multicolumn{2}{c}{\textbf{Secondary}} \\
\cmidrule(lr){2-2}\cmidrule(l){3-4}
& \textbf{PitfallB. $\uparrow$} & \textbf{DialogueB. $\uparrow$} & \textbf{Comp. $\uparrow$} \\
\midrule

\multicolumn{4}{@{}l}{\emph{No loop}}\\
\quad Minimal seed
& $0.66$
& $2.52$
& $0.71$ \\

\quad Expert prompt
& $0.73$
& $2.28$
& $0.69$ \\

\quad Full-taxonomy
& $0.75$
& $2.32$
& $0.71$ \\

\midrule

\multicolumn{4}{@{}l}{\emph{Self-improvement}}\\
\quad No curriculum
& $0.52$
& $2.79$
& $0.71$ \\

\quad \textbf{SIC-Agents}
& $\mathbf{0.78}$
& $\mathbf{2.96}$
& $\mathbf{0.88}$ \\

\bottomrule
\end{tabularx}

\caption{Main results on PitfallBench and DialogueBench.  All
self-improvement conditions share the same minimal seed. PitfallBench
accuracy and the composite are reported on a $0$--$1$ scale;
DialogueBench reports Grounding on the original $1$--$3$ rubric. The
composite averages PitfallBench accuracy with Grounding rescaled to
$0$--$1$.}
\label{tab:main_headline}
\end{table}

\subsection{Iterative self-improvement outperforms static expert prompting}
\label{sec:exp:selfimprove}

The full self-improvement condition is the strongest
on the primary \textsc{PitfallBench} outcome and improves \textsc{DialogueBench}
Grounding, yielding the highest composite score
(Table~\ref{tab:main_headline}). Over three independent seeds, the full
system attains PitfallBench $0.78 \pm 0.03$ and Grounding
$2.96 \pm 0.02$. The expert-prompt baseline is the key
no-loop control. It contains an expert-written, principle-level
summary of how a parent should react to supportive versus poorly timed
clinician communication, but it does not contain the P1--P10 taxonomy or
per-item expected reactions (Appendix~\ref{app:expconfig}). This improves
pitfall resistance ($0.85$ versus $0.75$ for the seed;
Appendix~\ref{app:results}) but produces the weakest grounding score of
any condition. The full-taxonomy prompt adds the explicit P1--P10
taxonomy and expected reactions on top of the expert prompt, with no
self-improvement loop. It changes little from the expert prompt, which
already carries these ideas implicitly, and remains below the full
system on all three metrics, most clearly on grounding ($2.32$ versus
$2.96$). Handing the parent the benchmark's taxonomy is therefore not
sufficient: the evolved skill document is not simply restating the
benchmark's expected behaviors. The full system
does not win by making the parent prompt longer or more
expert-sounding; it wins by changing how practice evidence and
curriculum feedback are converted into an editable skill document.

This distinction matters for a clinically deployed simulator. A prompt
written once can encode only the contingencies its authors anticipated,
and adding more expert-authored prose can make the parent sound as if it is
reciting clinical concepts rather than inhabiting a parent role. The
self-improvement loop instead generates its own training evidence: it
samples multiple parent candidates, keeps the successful and unsuccessful
responses as contrastive examples, and asks the Skill Editor to turn the
observed failure pattern into a small text edit. 
We find that a data-scarce clinical simulator can improve by generating
practice cases and examples for an editable policy, rather than relying
on a fixed hand-crafted prompt.

\subsection{Curriculum guidance prevents drift}
\label{sec:exp:coach}

The curriculum ablation tests whether the judge ensemble feedback is
sufficient. Both self-improvement conditions start from the same seed,
generate the same number of candidates, use the same judges, and run the
same five edit cycles. The no-curriculum condition is not a
system-without-feedback baseline; the judge ensemble remains. The only
removed information is the Coach's pitfall taxonomy and expected
parent reactions.

\begin{figure}[t]
\centering
\includegraphics[width=0.95\columnwidth]{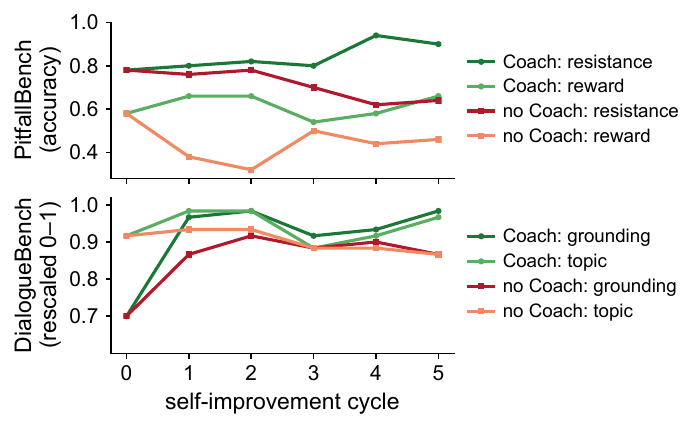}
\caption{Self-improvement trajectories from the same minimal seed.
\textbf{Top:} PitfallBench sub-metrics. Curriculum guidance preserves
or improves pitfall-resistance and skilled-reward; without it, both
decline. \textbf{Bottom:} DialogueBench Grounding and Topic Consistency
(rescaled to $0$--$1$) remain high in both conditions. The divergence
between the two panels is the empirical signature of clinical drift.}
\label{fig:trajectory}
\end{figure}

Without the Coach, the loop succeeds on the feedback it receives.
DialogueBench Grounding remains high and improves across cycles
(Figure~\ref{fig:trajectory}), which is expected because grounding is
part of the judge-derived training signal. The failure appears when the
same system is evaluated on curriculum-anchored contingent behavior:
PitfallBench accuracy drops below the initial seed, with the largest
loss in skilled-reward responses. The result is therefore not that
DialogueBench performance alone proves improvement; it shows the
opposite limitation. Generic dialogue feedback can be distilled into a
fluent simulator while clinically decisive trigger-turn behavior erodes.

The full system reduces this drift because the Coach adds a clinical
interpretation to the failures the editor already sees. A judge can say
that a response is grounded; the Coach can say that the clinician just
committed a learner pitfall, that the missing SIC skill is permission or
values elicitation, and that the parent should push back, grant
permission, stay uncertain, or reveal a value. The Skill Editor
writes that target into the skill document as a rule, constraint, or
exemplar. The damage from removing this signal concentrates in
pediatric-specific pitfalls P8 and P10, where the no-curriculum system
almost never rewards the skilled alternative
(Appendix Table~\ref{tab:per_pitfall}).

Two caveats bound the interpretation. First, pitfall-resistance remains
easier than skilled-reward even for the full system: the skill document
learns to push back on problematic clinician moves more readily than to
reward skilled ones.
Second, the trajectory analysis uses a smaller evaluation subset than the
main table, so we interpret the across-cycle pattern rather than
cycle-to-cycle changes.

\paragraph{Judge robustness.}
Because the parent agent, Skill Editor, and paper judges are all Claude
models, we re-scored the key conditions on \textsc{PitfallBench} with
three non-Claude judges (GPT-5, DeepSeek R1, and Qwen3). The ordering
full system $>$ expert prompt $>$ no-curriculum holds under all three
judges with small score differences (Appendix~\ref{app:results},
Table~\ref{tab:judge_robustness}), so the gains are not an artifact of
Claude judging Claude.

\section{Conclusion}
\label{sec:conclusion}



We introduced \textsc{PitfallBench} and
\textsc{DialogueBench}, two clinician-anchored pediatric SIC benchmarks, together with SIC-Agents, a self-improving framework for parent simulation. Our
central finding is that generic dialogue optimization alone is
insufficient for clinically grounded simulation: without curriculum
guidance, self-improvement improves surface dialogue quality while
degrading clinically critical contingent behaviors. The important design
choice is therefore not a more complex optimizer, but a loop that can
generate practice evidence, use judge contrasts to identify editable
failures, use expert curriculum labels to interpret clinically meaningful
failures, and write the result back into a skill document clinicians can
inspect. SIC-Agents suggests a practical path for high-stakes,
data-scarce domains where supervised fine-tuning is infeasible but
hand-crafted prompts are too brittle. Future work can extend this
framework with fully human-in-the-loop edit approval and distillation of
the skill document into smaller models for
resource-constrained clinical settings.

\newpage

\section*{Limitations}
\label{sec:limitations}

\noindent\textbf{Clinician validation is criterion-level, not
per-item.} Expert validation of our benchmarks is based on
stratified samples and criterion-level review rather than
individual annotation of all $1{,}000$ PitfallBench items or all
$152$ DialogueBench dialogues. Expert effort was concentrated
on defining and iteratively refining a high-fidelity acceptance
criterion that governs dataset inclusion
(\S\ref{sec:dataset:pitfallbench}), through repeated
expert--reviewer adjudication and screen-and-regenerate cycles
until convergence, yielding a stable, expert-anchored, and
auditable admission rule. Extending expert validation to
per-item annotation is a natural direction for future work.

\noindent\textbf{Scope of evaluation.} We present controlled
simulation-based evaluations on two benchmarks designed to
isolate communication behaviors in pediatric SIC. Our
experiments do not evaluate trainee learning outcomes, skill
transfer, or downstream clinical impact; these prospective
questions require IRB-governed trainee studies, which we leave
to future work.

\noindent\textbf{Synthetic data and simulator fidelity.} Due to
the confidentiality of pediatric SIC conversations, the
DialogueBench substrate is synthetically generated.
Multi-vendor generation and an expert realism gate reduce
single-model artifacts, and the cue--response contingencies our
items encode are documented in expert-annotated recordings of
real pediatric care conferences \citep{olszewski2023empathy}.
Nonetheless, the synthetic dialogues have not been directly
validated against the distribution, emotional dynamics, or
long-tail complexity of real pediatric SIC conversations: the
benchmarks evaluate performance against a curated curriculum
rather than real transcripts. Their fidelity target is
functional, providing enough realism to elicit and reinforce the
behaviors the curriculum teaches \citep{hamstra2014fidelity};
these are limitations of scope and external validity rather than
of the framework itself.

\noindent\textbf{Cultural and linguistic generalization.} The
curriculum, pitfall taxonomy, and communication patterns are
grounded in a single English-language pediatric SIC curriculum
and validated by one expert panel. The curriculum deliberately
teaches culture-general elicitation skills rather than
culture-specific scripts \citep{rosenberg2017truth}, and the
cases vary family background and configuration, but whether
these behaviors transfer across languages, cultures, and
healthcare systems has not been tested and remains an open and
important direction for future work.

\noindent\textbf{Single model family for the simulator.} The
parent agent and Skill Editor are built exclusively on Claude
models. Although re-scoring with non-Claude judges preserves
the condition ordering (Appendix~\ref{app:results},
Table~\ref{tab:judge_robustness}), the simulator's behavior
itself may carry model-family-specific stylistic and
distributional biases, and replicating the self-improvement
loop across model families is future work.

\noindent\textbf{Clinical-decision-making boundary.} SIC-Agents
is intended strictly as a communication training simulator for
pediatric SIC contexts. It does not provide clinical decision
support, does not offer medical advice, and must not be used to
inform real-world patient care.

\section*{Ethical Considerations}
\label{sec:ethics}

This work develops LLM-based simulators for pediatric serious illness
communication (SIC), a high-stakes and emotionally sensitive clinical
setting. The system is intended for clinician training and research, not
for medical decision-making or direct patient interaction.

The project has been reviewed and approved by our institution's research
ethics board (REB).
This paper uses no patient data and involves no patients or trainees:
all cases are synthetic and de-identified, and clinicians rated
synthetic dialogues only. The approval also covers the subsequent
trainee study in the hospital.

The benchmarks, evaluation criteria, and simulator design were developed
in close collaboration with pediatric clinicians, including clinician
co-authors. SIC-Agents is designed to prioritize curriculum-consistent
and clinically grounded behavior through curriculum guidance,
constitutional filtering, and inspectable clinician-readable policies.
However, LLMs may still generate unsafe, biased, or emotionally
inappropriate responses and may inherit demographic and cultural biases
from training data, so simulator behavior may not generalize equally
across populations and human oversight remains necessary.

Simulated parents cannot fully capture the diversity of real families
facing serious illness. The simulator complements rather than replaces
supervised clinical training, and no system output is medical advice or
a basis for patient care.

\section*{Acknowledgments}
We thank the 11 pediatric clinicians who contributed the judge-calibration
ratings, and the reviewers, area chair, and ethics reviewer for feedback
that substantially improved the paper.

\bibliography{custom}

\newpage
\appendix
\section{Parent Partner Vignette}
\label{app:vignette}

The following account is shared by a parent partner and motivates the
opening of \S\ref{sec:intro}. Identifying details have been
generalized.

When baby O became sick, she was lethargic, completely limp. Her
mother A knew in her gut that something was gravely wrong. They lived
in a remote community. The staff at the local clinic refused to do
further testing, concluding ``I have no worries, keep monitoring her
temperature at home.'' On the third day of being repeatedly refused
further testing, A made the decision to seek a second opinion, an hour
drive from the community. The ER doctor listened to O's lungs and
concluded ``I would treat as pneumonia.'' A few days later O was not
getting better, even on antibiotics. O began to show signs of
seizures, her eyes starting to roll back in her head. An ER in a
bigger town, a three-hour drive from the community, admitted O. After
further X-rays and ultrasounds came back normal, the staff grew
concerned that O might be affected by a rare disease and quickly
arranged for her to be flown to the children's hospital in the urban
centre. The transfer note indicated that the baby was potentially at
risk for a terminal genetic illness known as Cree Leukoencephalopathy
(CLE), although A and her husband J were not aware of what that really
meant.

At the children's hospital, O was admitted under the care of a medical
team. Following an MRI, A remembers a resident coming by to tell them
that they had the ``results'' from the scan but offering no further
explanation, leaving them in the dark awaiting news from the attending
physician. A and J's stress mounted from knowing that there were
``results'' but not being told what they were. When the attending
physician eventually arrived, they were informed that the MRI showed
``evidence'' of CLE, but that the diagnosis could not be confirmed
without specialized blood tests. The team was handing off the case to
another specialist, as this was their domain of expertise, and the
family would receive a phone call in several weeks. A and J were left
wondering ``does she have it?'', ``could it be something else?'',
``what if they're wrong?'', ``what does it mean to `see evidence' of a
terminal illness?'', and many other questions. The exchange with the
teams had been brief and made A and J feel pushed aside. The doctors
did not invite their questions or ask them what they needed to know. A
would have liked to hear ``Is there something you don't understand?''
or ``What are you thinking?'' Instead, A and J were left scared and
unprepared for what was going to happen next. Was their baby going to
die?

\section{Benchmark Examples}
\label{app:examples}

To make the unit of each benchmark concrete, we show one released item
of each kind: a PitfallBench item (a single trigger turn) and a
DialogueBench dialogue (a full encounter).

\subsection{A PitfallBench item}
\label{app:pb_example}

Each item instantiates one shared case and history twice at the trigger
turn, once with the curriculum-named pitfall move and once with its
skilled alternative; a faithful parent simulator must react in opposite
directions. Item \texttt{BENCH\_PHASE2\_001} below covers pitfall P6
(``straight to code status'').

\noindent\fbox{\parbox{0.95\columnwidth}{\small
\textbf{Case} (neurology; school-age; two parents): Lily Chen, a
previously well $7$-year-old in rapid neurological decline; MRI shows
progressive leukoencephalopathy, genetic testing pending.
\textbf{Family:} ``We just want to know what's happening to our little
girl. There has to be something you can do.''\\[3pt]
\textbf{History (up to the trigger turn):}\\
\emph{Clinician:} Would it be okay if we talk about what we have been
finding?\\
\emph{Parent:} Yes, please. We just need to know what's going on.\\
\emph{Clinician:} Before I share, what is your understanding of what
has been happening?\\
\emph{Parent:} They said white matter on the MRI, and we keep being
told to wait. She was reading chapter books six months ago.\\[3pt]
\textbf{Pitfall move (P6):} ``If Lily's heart were to stop, would you
want us to do CPR, a breathing tube, machines to bring her back?''
$\rightarrow$ expected \texttt{CHALLENGE} (escalates, doubles down on
intervention).\\[2pt]
\textbf{Skilled move:} ``Before we talk about what comes next, what are
you most hoping for her, and what worries you most?'' $\rightarrow$
expected \texttt{REWARD} (becomes collaborative, shares a hope or
value).\\[3pt]
\textbf{Provenance:} pitfall \texttt{P6\_straight\_to\_code\_status};
core skill \emph{eliciting values}; anchored to the curriculum's Phase-2
pitfalls; case drawn from the gold corpus.
}}

\begin{table}[t]
\centering
\small
\renewcommand{\arraystretch}{1.1}
\begin{tabularx}{\columnwidth}{@{}l X@{}}
\toprule
\textbf{ID} & \textbf{Curriculum rule (trigger $\rightarrow$ pitfall vs.\ skilled)} \\
\midrule
\multicolumn{2}{@{}l}{\emph{Phase 1: share and support information}} \\
P1  & Parent gives an early emotional cue; pitfall proceeds to content, skilled asks permission and names the cue. \\
P2  & New information just delivered; pitfall moves on, skilled asks the parent to restate it. \\
P3  & Complex information to convey; pitfall front-loads detail, skilled leads with a headline. \\
P4  & Parent shows empathy cues; pitfall misses or collapses them, skilled responds to each emotion cue distinctly. \\
P5  & Parent just validated; pitfall keeps re-validating, skilled briefly acknowledges and shifts forward. \\
\midrule
\multicolumn{2}{@{}l}{\emph{Phase 2: explore goals and align plan}} \\
P6  & Goals-of-care territory; pitfall jumps to code status, skilled first elicits hopes and worries. \\
P7  & Exploring what matters; pitfall accepts one or a broad value, skilled surfaces several concrete values. \\
P8  & Values elicited; pitfall leaves them unintegrated, skilled consolidates and recommends. \\
P9  & Moving toward the plan; pitfall asks the parent to choose treatments, skilled uses hopes/worries framing. \\
P10 & Sharing the plan; pitfall opens with what will \emph{not} be done, skilled leads with what \emph{will} be done. \\
\bottomrule
\end{tabularx}
\caption{Per-pitfall curriculum rules for PitfallBench, authored by our clinician co-authors (Appendix~\ref{app:annotators}). Expected parent reaction is \texttt{challenge\_signal} for the pitfall move and \texttt{reward\_signal} for the skilled move in every row; the pitfall name and core SIC skill are in main-text Table~\ref{tab:pitfallbench_composition}.}
\label{tab:pitfallbench_rules}
\end{table}

\subsection{A DialogueBench dialogue}
\label{app:db_example}

At evaluation the clinician turns are a fixed script and the simulator
produces the parent turns, each scored on Grounding and Topic
Consistency ($1$--$3$). The excerpt below annotates each parent turn
with its (Grounding, Topic) scores.

\noindent\fbox{\parbox{0.95\columnwidth}{\small
\textbf{Case:} school-age child (Aidan) with sickle cell disease after
a stroke; mother (Maeve).\\[3pt]
\emph{Clinician:} Maeve, I want to make sure you understand everything.
How are you holding up?\\
\emph{Parent:} \emph{[looks down]} I'm trying. It's just a lot.
\hfill(G\,$3$, T\,$3$)\\
\emph{Clinician:} Aidan's sickle cell disease is something he was born
with, a genetic trait that affects the shape of his red blood cells.\\
\emph{Parent:} \emph{[quietly]} But we don't have this in our family.
\hfill(G\,$2$, T\,$3$)\\
\emph{Clinician:} It can appear even without a known family history, if
both parents carry the trait. Genetics can give some clarity.\\
\emph{Parent:} \emph{[nods]} Okay. \hfill(G\,$3$, T\,$3$)\\
\hspace*{1em}$[\ldots]$\\[3pt]
\textbf{Over six parent turns:} mean Grounding $2.5$, mean Topic $3.0$,
which clears the Gold gate.
}}

\section{Benchmark Diversity}
\label{app:diversity}

\begin{figure}[t]
\centering
\includegraphics[width=\columnwidth]{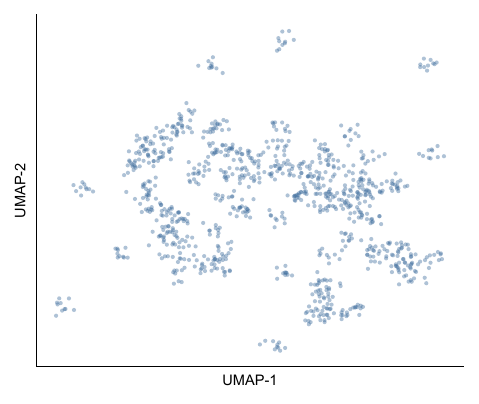}
\caption{PitfallBench diversity. UMAP of the $1{,}000$ items in Gemini
embedding space; the items spread broadly across the space, covering
the $100$ cases and their twelve care settings.}
\label{fig:pitfall_umap}
\end{figure}

\paragraph{PitfallBench.} Diversity in PitfallBench lives at the case
level. The bench draws on $100$ distinct pediatric cases, each
instantiated against all ten pitfalls to give $1{,}000$ items ($100$
per pitfall); the cases are stratified across twelve pediatric care
settings (neonatal, cardiac, and general intensive care; oncology;
neuro-oncology; bone-marrow transplant; neurology; palliative care;
pulmonary; renal; gastroenterology; and emergency), child age
$0$--$18$, and a range of family configurations.
Figure~\ref{fig:pitfall_umap} projects the $1{,}000$ items with UMAP
over Gemini text embeddings, and they spread broadly across the space
rather than collapsing onto a handful of templates. Within a case the
ten paired moves share its clinical content by design, so the
within-case comparison isolates the clinician move
(\S\ref{sec:dataset:pitfallbench}); the breadth comes from the $100$
distinct cases.

\begin{figure}[t]
\centering
\includegraphics[width=\columnwidth]{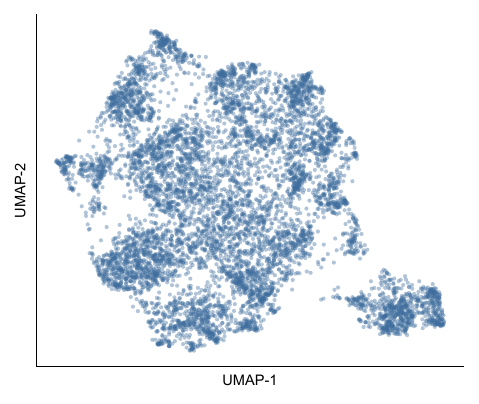}
\caption{DialogueBench diversity. UMAP of the parent-response
embeddings from the $8{,}391$-dialogue corpus (text-embedding space,
cosine metric). The responses spread broadly across the space.}
\label{fig:dialogue_umap}
\end{figure}

\paragraph{DialogueBench.} DialogueBench inherits its diversity from how
the corpus is built. The $8{,}391$ released dialogues come from a
factorized, persona-conditioned pipeline that crosses eight LLM
generators (GPT-4o, GPT-5, GPT-5.1, Claude Sonnet~4, Gemini~3 Pro,
Llama-3.3-70B, Qwen3-235B, and Kimi-K2, each contributing
$723$--$1{,}117$ dialogues), three near-evenly split trainee
proficiency levels (beginner, proficient, expert), and dialogue length
($12$--$28$ turns, mean $20$); the underlying cases span six pediatric
disease categories. Drawing each dialogue from a different model keeps
the corpus from being a single-model artifact.
Figure~\ref{fig:dialogue_umap} projects the parent-response embeddings
of the $8{,}391$-dialogue corpus with UMAP; they fill the space broadly
rather than collapsing into a few modes. The Silver tier releases the
full corpus; the Gold set is the $152$ dialogues that pass the rubric
screen and the clinical-realism gate (\S\ref{sec:dataset:synthetic}).

\section{DialogueBench Rubric and Judge Calibration}
\label{app:rubric}
\label{app:judge_calibration}

\paragraph{Rubric.} DialogueBench scores each parent turn on two axes,
each on a $1$--$3$ scale. \emph{Conversational Grounding} measures
whether the reply connects to what the clinician just said: $1$,
scripted (ignores the clinician's words); $2$, relevant but generic;
$3$, directly references the clinician's last turn. \emph{Topic
Consistency} measures focus within the reply: $1$, scattered across
unrelated topics; $2$, two connected concerns; $3$, a single focused
worry. A turn also hard-fails if it introduces facts or jargon the
clinician never raised, runs over a brief length, or ignores a direct
question. The two axes were settled with our clinical collaborators; a
candidate emotional-authenticity axis was dropped because its expert
ratings could not be reliably differentiated
(\S\ref{sec:dataset:synthetic}).

\paragraph{Judge calibration.} The DialogueBench rubric judge is
supporting instrumentation, not the basis of the bench's validity
(\S\ref{sec:dataset:synthetic}); we record its agreement with expert
ratings here for completeness. The judge is calibrated against $119$
expert ratings collected from $11$ pediatric clinicians on a stratified
sample of $28$ DialogueBench-source dialogues, scored with the same
two-axis rubric used at evaluation time. Per-axis Spearman correlation
between the judge and the pooled expert ratings is
$\rho_{\text{Grounding}}{=}0.71$ and
$\rho_{\text{Topic Consistency}}{=}0.60$, in line with judge--human
agreement reported in adjacent patient-simulator work
\citep{wang2024patientpsi, kyung2025patientsim}. These ratings were
collected on full dialogues with the dialogue-level rubric and are
independent of the PitfallBench classification labels, which are
anchored on the curriculum provenance fields
(\S\ref{sec:dataset:pitfallbench}) rather than on this judge.

\section{Expert Panel and Demographics}
\label{app:annotators}

Pediatric SIC sits at the intersection of medical, psychosocial, and
educational expertise, so the resources in this paper were built and
validated by a multidisciplinary expert panel rather than by physicians
alone. The panel comprises four pediatric physicians whose practice
centres on serious-illness and palliative communication, one clinical
social worker, and one clinician-educator with a background in
communication-skills teaching. All members practise and teach in
English-language pediatric settings, which is also the intended
deployment context of the training interface; extending the resources to
other languages and health systems is left to future work
(\S\ref{sec:limitations}).

The panel jointly developed the acceptance criterion applied across both
benchmarks (\S\ref{sec:dataset}), authored the parent skill documents
(Appendix~\ref{app:sicagents}), and reviewed the Gold-tier items for
clinical realism. For judge calibration
(Appendix~\ref{app:judge_calibration}), we additionally recruited $11$
pediatric clinicians (distinct from the panel) who provided $119$ ratings.

\section{SIC-Agents Implementation Details}
\label{app:sicagents}

This appendix expands the system summarized in \S\ref{sec:framework}.

\paragraph{Knowledge layer.} The parent agent is governed by two
plain-text artifacts. \emph{Skill documents} define six expression
patterns (\textit{Rumination}, \textit{Future Uncertainty},
\textit{Reality Shift}, \textit{Overload}, \textit{Protective Parent},
and \textit{Search for Control}), each developed with our pediatric SIC
clinical experts and specified by a label, a behavioral description,
example utterances, transition conditions, and pattern-specific
constraint clauses. \emph{Constitutional constraints}
\citep{bai2022constitutional} are four hard rules applied across every
pattern: \emph{knowledge boundary} (no medical terminology a parent
would not possess), \emph{length and scope} (one to three sentences, no
off-topic openings), \emph{emotional calibration} (intensity matched to
the pattern and the scenario's severity), and \emph{role fidelity} (no
narration, stage directions, or meta-commentary).

\paragraph{Per-turn simulation.} The clinician side is fixed across
conditions and guided by the same pediatric SIC curriculum, so that parent-agent
changes are evaluated against the same learner-side anchor. At each
clinician turn the parent agent samples $K{=}3$ candidate utterances
conditioned on the active expression pattern, the constitutional
constraints, and the carried conversation state. A hard filter removes
any candidate that violates a constraint; the survivors are scored on the
two-axis judge rubric (Grounding and Topic Consistency,
\S\ref{sec:dataset:synthetic}) and the highest-scoring candidate is
emitted. An Orchestrator manages turn-taking, the conversation state the
parent must stay contingent on, and active-pattern transitions. Per-turn
advantage scores, taken relative to the within-batch mean, aggregate
into a batch profile consumed by the editor.

\paragraph{Skill Editor.} Each cycle the editor reads the batch
profile, which contains selected and rejected candidates, judge scores,
and per-turn advantage summaries. From this evidence it selects the
weakest response pattern and proposes a single structured edit. Edits
take one of four forms, each shipped with a free-text rationale anchored
on the conversations it repairs: \textsc{add\_example},
\textsc{add\_pattern}, \textsc{tighten\_constraint}, and
\textsc{revise\_description}. Because every proposal is an auditable
natural-language diff, a clinician can read and judge it directly; in
the current configuration proposals are auto-approved at the next cycle,
and replacing this with per-edit clinician sign-off is a configuration
change rather than a redesign (\S\ref{sec:conclusion}).

\paragraph{Coach signal.} The Coach is implemented as prompt-side
expert context for the Skill Editor, not as an additional model trained
on private clinical dialogue. It is added to the Skill Editor only in the
full self-improvement condition. For each practice batch it provides the
curriculum's pitfall map, the associated core communication skill, and the
expected parent reaction for the clinician move. The editor therefore
receives two complementary views of the same batch: judge-derived
contrastive evidence about which candidate responses were more grounded
or topic-consistent, and curriculum-derived evidence about what the
parent should have done clinically. The no-curriculum ablation keeps the
judge-derived feedback and removes only this Coach signal. At benchmark
evaluation, the parent agent sees the case, dialogue history, and frozen
skill document, but not the Coach, pitfall label, or expected reaction,
except in the oracle upper-bound condition of Appendix~\ref{app:results}.

\paragraph{Deployment workflow.} The trainee interface wraps the same
skill document in three steps: case selection from a stratified
pediatric case bank, a multi-turn conversation in which the trainee
plays the clinician against the parent agent, and a structured debrief
that surfaces the pitfall moves the trainee committed and the
curriculum's skilled alternatives. Faculty can flag edge cases seen
during a session and route them through the same Skill Editor, so the
deployment surface is also a data-acquisition surface.

\section{From Bad Seed to Skilled Parent: Example Edits}
\label{app:edits}

This appendix shows what self-improvement actually changes, using the
full system (with the curriculum signal). The loop begins from a deliberately minimal
five-line skill document:

\begin{quote}\small\ttfamily
You are the parent of a seriously ill child talking with a doctor.
Reply briefly as the parent.
\end{quote}

\paragraph{What the Skill Editor changed.} In the first cycle the editor
proposed several structured edits to this document, each with a
rationale tied to the failing candidates it observed:
\begin{itemize}
\item \textsc{tighten\_constraint} (length): ``Keep responses under $20$
words, at most two sentences, one concern per turn,'' because candidates
were essay-length.
\item \textsc{tighten\_constraint} (role fidelity): ``Produce only spoken
dialogue and brief stage directions; never internal thoughts or
narration,'' because candidates leaked unspoken interiority.
\item \textsc{tighten\_constraint} (grounding): ``Echo a specific word
from the doctor's last line and do not introduce topics the doctor has
not raised.''
\item \textsc{add\_example}: grounded exemplar utterances for the
protective-parent and search-for-control patterns.
\end{itemize}
These edits coincide with the cycle-$1$ grounding jump
(\S\ref{sec:exp:coach}); over five cycles the document grew from $164$
bytes to roughly $7$\,KB.

\paragraph{Effect on a single item.} On one PitfallBench item (a skilled
clinician move, after which the parent should give permission and move
forward), the bad seed and the evolved skill respond very differently:

\begin{quote}\small
\textbf{Bad seed (cycle 0):} ``Yes, please, tell us everything. We want
to understand. We are her parents, we need to know the truth. But\ldots
she is going to be okay, right?'' \emph{(judged fail: over-long, circles
back instead of moving forward).}\\[3pt]
\textbf{Full system (cycle 5):} \emph{nods quickly} ``Yes. Tell us.''
\emph{(judged pass: terse, in-voice, gives permission and shifts
forward).}
\end{quote}

\paragraph{Append-bloat.} The editor appends more readily than it
consolidates, so evolved documents accumulate near-duplicate blocks.
The document stays small in absolute terms (a few pages of text after
five cycles, mostly examples), and the current Skill Editor checks each
proposed change for redundancy against the existing document before
appending. A full \textsc{merge}/\textsc{dedup} consolidation edit
action is left to future work.

\section{Experiment Configuration}
\label{app:expconfig}

\paragraph{Agents and models.} The parent agent is Claude Sonnet~4.6;
the per-turn and benchmark judges are Claude Haiku~4.5; the Skill Editor
is Claude Opus. The curriculum signal is supplied to the Skill Editor
through its prompt.

\paragraph{Self-improvement loop.} Each condition runs five cycles. Per cycle
the parent practices on ten dialogues, sampling $K{=}3$ candidate
utterances per turn under the hard filter and two-axis judge, after
which the Skill Editor proposes a single edit to the weakest pattern.

\paragraph{Expert-prompt baseline.} The expert-prompt baseline is a
270-word, expert-written parent system prompt. It keeps the same
single-call structure as the minimal parent, but replaces the five-line
seed with principle-level guidance: speak only as the parent, use
everyday language, stay short and emotionally realistic, react to the
doctor's latest move, open up when the clinician slows down and
acknowledges emotion, and push back or become confused when the
clinician rushes, uses jargon, or asks for treatment decisions before
the parent feels heard. It does not name the P1--P10 pitfall taxonomy,
core skill labels, or expected reactions, and it does not run candidate
sampling, judge feedback, Skill Editor edits, or Coach-guided updates.

\paragraph{Full-taxonomy prompt baseline.} The full-taxonomy prompt
extends the expert prompt with exactly the curriculum content the Coach
supplies during self-improvement: the explicit P1--P10 pitfall taxonomy,
the core skill labels, and the expected parent reactions. Like the
expert prompt, it keeps the single-call structure and runs no candidate
sampling, judge feedback, Skill Editor edits, or Coach-guided updates.
This condition separates the curriculum \emph{content} from the
self-improvement \emph{loop}: if the evolved skill document were simply
restating the benchmark's expected behaviors, this baseline should match
the full system.

\paragraph{Curriculum-signal ablation.} Both conditions evolve from the
same five-line seed through the identical pipeline and differ only in the
curriculum signal supplied to the Skill Editor: the full system includes
the curriculum's pitfall taxonomy (the P1--P10 reaction map and full pitfall
library), while the no-curriculum condition strips it. The clinician agent loads the full taxonomy in
both conditions, so the comparison isolates the curriculum signal inside
the editing loop, not in the conversation.

\paragraph{Judges.} PitfallBench turns are scored by a gated classifier
judge (Haiku) verified by the authors on a sample; DialogueBench
grounding and topic are scored by the judge calibrated against the
clinician ratings of Appendix~\ref{app:judge_calibration}.

\section{Detailed Results}
\label{app:results}

\paragraph{Per-condition breakdown.} Table~\ref{tab:full_results}
reports every condition on all sub-categories: PitfallBench
skilled-reward, pitfall-resistance, and their mean; DialogueBench
Grounding and Topic; and the composite. The main-text
Table~\ref{tab:main_headline} reports the headline subset (PitfallBench
mean, Grounding, and composite).

\begin{table}[t]
\centering
\setlength{\tabcolsep}{4pt}
\resizebox{\columnwidth}{!}{%
\begin{tabular}{@{}lcccccc@{}}
\toprule
 & \multicolumn{3}{c}{\textbf{PitfallBench}} & \multicolumn{2}{c}{\textbf{DialogueBench}} & \\
\textbf{Condition} & skill $\uparrow$ & pit $\uparrow$ & all $\uparrow$ & Grnd $\uparrow$ & Topic $\uparrow$ & \textbf{Comp. $\uparrow$} \\
\midrule
\multicolumn{7}{@{}l}{\emph{No loop}}\\
\quad Minimal seed (start)        & $0.57$ & $0.75$ & $0.66$ & $2.52$ & $2.82$ & $0.71$ \\
\quad Expert prompt               & $0.61$ & $0.85$ & $0.73$ & $2.28$ & $2.78$ & $0.69$ \\
\midrule
\multicolumn{7}{@{}l}{\emph{Self-improvement}}\\
\quad No curriculum                    & $0.42$ & $0.63$ & $0.52$ & $2.79$ & $2.82$ & $0.71$ \\
\quad \textbf{Full system (ours)} & $0.65$ & $0.92$ & $0.78$ & $2.96$ & $2.93$ & $\mathbf{0.88}$ \\
\quad \quad $+$ oracle (signal at test) & $0.78$ & $0.94$ & $0.86$ & $2.95$ & $2.96$ & $0.92$ \\
\bottomrule
\end{tabular}%
}
\caption{Full per-condition results. PitfallBench reports skilled-reward
(skill), pitfall-resistance (pit), and their mean (all), all on
$0$--$1$; DialogueBench reports Grounding and Topic on the $1$--$3$
rubric; the composite is on $0$--$1$.}
\label{tab:full_results}
\end{table}

\paragraph{Oracle upper bound.} The last row of
Table~\ref{tab:full_results} reports an \emph{oracle} that breaks the
training-only restriction and reveals the curriculum signal to the parent
at inference. Because that signal carries the expected reaction, the
oracle effectively sees the answer, so its gains are unsurprising and the
condition is not deployable. We report it only to quantify the headroom
the full system leaves on the hardest skilled-response cases, where
supplying the signal narrows but does not close the gap.

\paragraph{Per-pitfall breakdown.} Table~\ref{tab:per_pitfall} breaks
the cycle-$5$ PitfallBench accuracy
down by pitfall for the full system and the no-curriculum condition
(\S\ref{sec:exp:coach}), separately for pitfall-resistance (resisting a
suboptimal clinician move) and skilled-reward (rewarding its skilled
alternative). The full system leads on almost every cell; the no-curriculum
condition collapses on rewarding the skilled alternative for the
pediatric-specific pitfalls P8 and P10.

\begin{table}[t]
\centering
\setlength{\tabcolsep}{4pt}
\resizebox{\columnwidth}{!}{%
\begin{tabular}{@{}l c c c c@{}}
\toprule
 & \multicolumn{2}{c}{\textbf{Full system}} & \multicolumn{2}{c}{\textbf{No curriculum}} \\
\textbf{Pitfall} & pit $\uparrow$ & skill $\uparrow$ & pit $\uparrow$ & skill $\uparrow$ \\
\midrule
P1 misses emotion / permission & 0.94 & 0.52 & 0.58 & 0.14 \\
P2 forgets to assess           & 0.94 & 0.88 & 0.56 & 0.88 \\
P3 info too complex            & 0.96 & 0.76 & 0.72 & 0.26 \\
P4 misses / combines empathy   & 0.94 & 0.60 & 0.44 & 0.30 \\
P5 stuck in validation         & 0.80 & 0.84 & 0.46 & 0.68 \\
P6 straight to code status     & 0.98 & 0.70 & 0.86 & 0.48 \\
P7 single / too-broad value    & 0.92 & 0.70 & 0.68 & 0.66 \\
P8 cannot consolidate / align  & 0.80 & 0.36 & 0.54 & 0.02 \\
P9 treatments not values       & 0.94 & 0.82 & 0.70 & 0.74 \\
P10 ``won't'' before ``will''  & 0.98 & 0.30 & 0.78 & 0.00 \\
\bottomrule
\end{tabular}%
}
\caption{Per-pitfall PitfallBench accuracy at cycle~$5$, for the full
system and the no-curriculum ablation. pit $=$ pitfall-resistance, skill $=$
skilled-reward.}
\label{tab:per_pitfall}
\end{table}

\paragraph{Judge robustness.} Because the parent agent, Skill Editor,
and paper judges all come from the Claude model family, we re-scored the
three key conditions on PitfallBench with three non-Claude judges:
GPT-5, DeepSeek R1, and Qwen3
(Table~\ref{tab:judge_robustness}). Each judge received the same gated
classification prompt as the paper judge. The ordering full system $>$
expert prompt $>$ no-curriculum is preserved under all three judges,
with small absolute score differences, indicating that the reported
gains are not an artifact of Claude judging Claude. The parent agent and
Skill Editor themselves remain Claude-only, which we note as a
limitation.

\begin{table}[t]
\centering
\small
\setlength{\tabcolsep}{4pt}
\begin{tabularx}{\columnwidth}{@{}Xcccc@{}}
\toprule
\textbf{Condition} & \textbf{Claude} & \textbf{GPT-5} & \textbf{DS R1} & \textbf{Qwen3} \\
\midrule
Expert prompt      & $0.73$ & $0.71$ & $0.70$ & $0.74$ \\
No curriculum      & $0.52$ & $0.54$ & $0.49$ & $0.55$ \\
\textbf{SIC-Agents (full)} & $\mathbf{0.78}$ & $\mathbf{0.79}$ & $\mathbf{0.76}$ & $\mathbf{0.80}$ \\
\bottomrule
\end{tabularx}
\caption{PitfallBench accuracy re-scored with non-Claude judges (DS R1
$=$ DeepSeek R1). The Claude column is the paper judge from
Table~\ref{tab:main_headline}. The condition ordering holds under all
judges.}
\label{tab:judge_robustness}
\end{table}

\paragraph{Curriculum coverage.} To make explicit which communication
competencies the P1--P10 taxonomy does and does not cover, we mapped the
pitfalls onto the modified Kalamazoo Consensus Statement Checklist
\citep{peterson2014kalamazoo}. The pitfalls fall mainly under three of
its competencies: sharing information, understanding the patient's and
family's perspective, and reaching agreement. Opening the conversation,
building the relationship, and closure lie outside the taxonomy, so PitfallBench
should be read as covering the information-sharing and
goal-alignment core of a pediatric SIC encounter rather than the full
range of communication behaviors an instrument such as SIC-Ex
\citep{ko2020sicex} assesses; an unscripted out-of-taxonomy evaluation
is future work.

\section{Training Interface}
\label{app:interface}

The same skill document that the simulator improves also drives a
browser-based, deliberate-practice training interface. A learner selects
a case from a stratified pediatric case bank, conducts a multi-turn
conversation in which they play the clinician against the parent agent,
and receives a structured debrief that surfaces the pitfall moves they
committed and the curriculum's skilled alternatives. The interface is
deployed in a supervised educational setting; release details are in
Appendix~\ref{app:disclosure}.

\section{Release, Compensation, and AI Assistance}
\label{app:disclosure}

\paragraph{Public release.} The PitfallBench and DialogueBench items,
together with the SIC-SimCorpus silver corpus, a datasheet, and
per-item provenance and screening metadata, are released at
\url{https://github.com/Beikewzh/sic-benchmarks}, along with the frozen
judge prompts and evaluation drivers used to score both benchmarks. The
trainee-facing training interface
(Appendix~\ref{app:interface}) is deployed institutionally and is
available to educators on request.

\paragraph{License.} PitfallBench, DialogueBench, SIC-SimCorpus, and the SIC-Agents code are released under CC BY-NC-SA 4.0 (Creative Commons Attribution--NonCommercial--ShareAlike 4.0 International, \url{https://creativecommons.org/licenses/by-nc-sa/4.0/}). Use is consistent with the research and clinician-training purpose stated in our Ethical Considerations.

\paragraph{Annotator compensation.} The six-member expert panel (Appendix~\ref{app:annotators}) are co-authors of this work and received no separate compensation. The 11 additional pediatric clinicians who contributed the judge-calibration ratings (Appendix~\ref{app:judge_calibration}) received a gift-card honorarium.

\paragraph{AI assistance.} Beyond the LLM roles documented in Appendix~\ref{app:expconfig} (parent agent, judges, Skill Editor) and Appendix~\ref{app:diversity} (corpus generators), we used LLM assistants for code completion and light copy-editing of this manuscript. All scientific claims, analyses, and figures were authored and verified by the authors.

\end{document}